\documentclass[12pt]{elsarticle}

\usepackage{amssymb}
\usepackage{amsthm}
\usepackage{mathtools}
\usepackage{color}
\usepackage{booktabs} 
\usepackage{algorithmic}
\usepackage{algorithm}
\usepackage{diagbox}
\usepackage{ulem}
\usepackage{subfigure}
\theoremstyle{plain}
\newtheorem{theorem}{Theorem}[section]

\theoremstyle{definition}

\newtheorem{example}[theorem]{Example}
\theoremstyle{remark}

\newcommand{\bd}[1]{\boldsymbol{#1}}
\newcommand{\mb}[1]{\mathbb{#1}}
\newcommand{\mc}[1]{\mathcal{#1}}
\newcommand{\nm}[2]{\|\,#1\,\|_{#2}}

\journal{arXiv}

\begin{document}

\begin{frontmatter}



\title{GAS: A Gaussian Mixture Distribution-Based Adaptive Sampling Method for PINNs}

\author[inst1,inst2]{Yuling Jiao}
\ead{yulingjiaomath@whu.edu.cn}
\author[inst1]{Di Li}
\ead{lidi.math@whu.edu.cn}
\author[inst1,inst2]{Xiliang Lu}
\ead{xllv.math@whu.edu.cn}	
\author[inst1,inst2]{Jerry Zhijian Yang}
\ead{zjyang.math@whu.edu.cn}
\author[inst1,inst2]{Cheng Yuan}
\ead{yuancheng@whu.edu.cn}

\affiliation[inst1]{organization={School of Mathematics and Statistics, Wuhan University},
            city={Wuhan},
            postcode={430072},
            country={China}}

\affiliation[inst2]{organization={Hubei Key Laboratory of Computational Science, Wuhan University},
            city={Wuhan},
            postcode={430072},
            country={China}}

\begin{abstract}
With the recent study of deep learning in scientific computation, the Physics-Informed Neural Networks (PINNs) method has drawn widespread attention for solving Partial Differential Equations (PDEs). Compared to traditional methods, PINNs can efficiently handle high-dimensional problems, but the accuracy is relatively low, especially for highly irregular problems. Inspired by the idea of adaptive finite element methods and incremental learning, we propose GAS, a Gaussian mixture distribution-based adaptive sampling method for PINNs. During the training procedure, GAS uses the current residual information to generate a Gaussian mixture distribution for the sampling of additional points, which are then trained together with historical data to speed up the convergence of the loss and achieve higher accuracy. Several numerical simulations on 2D and 10D problems show that GAS is a promising method that achieves state-of-the-art accuracy among deep solvers, while being comparable with traditional numerical solvers.
\end{abstract}




\end{frontmatter}


\section{Introduction}
\label{sec:Introduction}

In recent years, solving partial different equations(PDEs) with deep learning methods has been widely studied, e.g., the Physics-Informed Neural Networks (PINNs)~\cite{raissi2019physics}, the Deep Ritz Method (DRM)~\cite{Weinan17} and the Weak Adversarial Neural Networks (WAN)~\cite{Zang20}. While Both DRM and WAN use the variational forms of the PDEs, PINNs solves PDEs by a direct minimization of the square residuals, which makes the method more flexible and easier to be formulated to general problems. By reformulating the original PDE problems into a optimization problems, The physics-informed neural networks \cite{raissi2019physics, karniadakis2021physics, sirignano2018dgm, pang2019fpinns} can easily approximate the solution of PDE with a deep neural network (DNN) function space, through minimizing the corresponding loss term which is defined as the integral of residuals. In practice, this integral is approximated by using Monte Carlo (MC) methods with finite points, which are usually sampled according to a uniform distribution on the computational domain. In contrast to classical computational methods, where the main concern is the approximation error, one needs to balance the approximation error and the generalization error for the neural network solvers, where the approximation error mainly originates from the modeling capability of the neural network, while the generalization error is mainly related to the discretization of loss with random samples. While a uniform random sampling strategy is simple to implement, the low regularity regions of the solution may not be taken consideration enough, which makes PINNs be inefficient and even inaccurate, especially when the problems is high dimensional.

To handel the singularity in solution, several sampling strategies have been developed to improve the efficiency and accuracy of PINNs. In \cite{lu2021deepxde}, a simple residual-based adaptive refinement method RAR was proposed to iteratively add new points with the largest residuals in the training set. Later in \cite{wu2023comprehensive}, the authors give a comprehensive study of non-adaptive and residual-based adaptive sampling methods, with an extension of the RAR method to more general residual-based adaptive distribution methods (RAD and RAD-G), in which the added points are sampled from a combination of residual-induced density and uniform distribution. Meanwhile, motivated by the idea of self-paced learning, \cite{gu2021selectnet} proposed the SelectNet to adjust the weight of different samples in the MC integral of residuals, which allows the network to learn the most important points first. Apart from these, by reformulating the loss functional with the idea of importance sampling \cite{katharopoulos2018not, nabian2021efficient}, the authors developed a KRnet-based method DAS to approximate the proposed density \cite{tang2021das}, which proves to be an effective way to capture the singularity of solution. More recently, inspired by the idea of adaptive finite element scheme, researchers studies a failure-informed adaptive sampling method FI-PINNs \cite{gao2022failure, gao2023failure}, in which new points are added to better approximate the failure probability of PINNs. With the approximation of proposal density in the importance sampling of failure probability by Gaussians or Subset simulation, FI-PINNs shows a promising prospects in dealing with multi-peak and high dimensional problems.

In this paper, motivated by the concept of continual learning\cite{hadsell2020embracing} and the key idea in adaptive FEM that one should refine meshes according to some posterior indicator\cite{nochetto2009theory}, we proposed GAS, a Gaussian mixture distribution-based adaptive sampling strategy as a region refinement method. Different from the DAS and FI-PINNs methods in which the adaptive sampling is aimed at the approximation of some proposed density in importance sampling, we define the distribution of added data by maximizing the risk with the new density. The whole procedure of GAS can be viewed as an alternate iteration for solving a min-max problem. Furthermore, the model parameters in the Gaussian mixture model is calculated via Laplace approximation, rather than the density estimation strategy used in \cite{gao2022failure}. As a result, we select points where the residual is relatively bigger as means of Gaussians, and the gradient of residual at these points are used to construct the covariances respectively. After that, additional points sampled according to this mixture Gaussians will be added to the old training set, followed by the next training round of PINNs. With such an easy while effective incremental learning strategy, we successfully solved several highly irregular problems from 2 dimension to 10 dimension by PINNs, and our method achieve the state of the art in the field of adaptive sampling for PINNs. Furthermore, the min-max formulation together with the Laplace approximation techniques makes GAS an explainable adaptive sampling strategy, which also provides a theoretical framework for further error analysis. 


The rest of this paper is organized as follows. In Section 2, a brief overview of PINNs will be presented. The GAS method is then constructed in section 3 and we demonstrate the efficiency of our method with numerical experiments in section 4. In Section 5 the main conclusion will be given.

\section{Preliminaries of PINNs}
Let $\Omega \in \mb{R}^d$ be a bounded spatial domain and $\bd{x} \in \mb{R}^d$. The PDE problem is stated as: find $u(\bd{x}) \in F:\mb{R}^d \mapsto R$ where $F$ is a proper function space defined on $\Omega$, such that:
\begin{equation}\label{eq:problem}
  \left\{
  \begin{array}{ll}
  \mc{L}u(\bd{x})=s(\bd{x}), & \forall \bd{x} \in \Omega, \\[1.5ex]
  \mc{B}u(\bd{x})=g(\bd{x}), & \forall \bd{x} \in \partial \Omega. \\[1.5ex]
  \end{array}
\right.
\end{equation}

Here $\mc{L}$ and $\mc{B}$ are operators defined in the domain and on the boundary respectively. Let $u(\bd{x}\,; \Theta)$ be a neural network with parameters $\Theta$. In the framework of PINNs, the goal is to use $u(\bd{x}\,;\Theta)$ to approximate the solution $u(\bd{x})$ through optimizing the following loss function:

\begin{equation}\label{eq:loss}
\begin{aligned}
J(u(\bd{x} \, ; \Theta)) &= \nm{r(\bd{x} \, ; \Theta)}{2,\Omega}^2 + \gamma \nm{b(\bd{x} \, ; \Theta)}{2,\partial \Omega}^2 \\
&= J_r(u(\bd{x}\,;\Theta)) + \gamma J_b(u(\bd{x} \,;\Theta)),
\end{aligned}
\end{equation}
where $r(\bd{x} \,;\Theta) = \mc{L}u(\bd{x}\,;\Theta) - s(\bd{x})$, $b(\bd{x}\,;\Theta) = \mc{B}u(\bd{x}\,;\Theta)-g(\bd{x})$, and $\gamma > 0$ is a penalty parameter. In practice, the loss functional (\ref{eq:loss}) is usually discretized numerically by:

\begin{equation}\label{dis_loss}
J_N(u(\bd{x} \, ; \Theta)) = \nm{r(\bd{x} \, ; \Theta)}{N_r, S_\Omega}^2 + \hat{\gamma} \nm{b(\bd{x} \, ; \Theta)}{N_b,S_{\partial \Omega}}^2,
\end{equation}
where $S_\Omega = \{\bd{x}^{(i)}_\Omega\}_{i=1}^{N_r}$ and $S_{\partial \Omega} = \{ \bd{x}_{\partial \Omega}^{(i)} \}_{i=1}^{N_b}$ are two sets of uniformly distributed collocation points, and
\begin{equation*}
\begin{aligned}
  \nm{u(\bd{x} \, ; \Theta)}{N_r, S_\Omega} = (\frac{1}{N_r}\sum_{i=1}^{N_r}u^2(\bd{x}_{\Omega}^{(i)}))^{\frac{1}{2}}, \\ \nm{u(\bd{x} \, ; \Theta)}{N_r, S_{\partial \Omega}} = (\frac{1}{N_b}\sum_{i=1}^{N_b}u^2(\bd{x}_{\partial\Omega}^{(i)}))^{\frac{1}{2}}
\end{aligned}
\end{equation*}
are the empirical risks inside and on the boundary of $\Omega$. Defining the following two minimizers
\begin{equation}
\begin{array}{ll}
u(\bd{x} \, ; \Theta^{*}) = \arg \min_{\Theta} J(u(\bd{x} \,; \Theta)) \\
u(\bd{x} \, ; \Theta_N^{*}) = \arg \min_{\Theta} J_N(u(\bd{x} \,; \Theta)),
\end{array}\label{E-risk}
\end{equation}
we can decompose the error in PINNs as:
\begin{equation}
\begin{aligned}
\mb{E}(\nm{u(\bd{x}\,;\Theta_N^{*}) - u(\bd{x})}{\Omega}) &\leq \mb{E}(\nm{u(\bd{x}\,;\Theta^{*}) - u(\bd{x})}{\Omega}) \\
&+ \mb{E}(\nm{u(\bd{x}\,;\Theta_N^{*}) - u(\bd{x}\,;\Theta^{*})}{\Omega}).
\end{aligned}
\end{equation}
While the first term describes the approximation error related to the expressing capability of neural networks, the second term consists of the optimization error and the statistical error due to random sampling in the MC procedure. In this work, we focus on the reduction of statistical error by seeking for a better estimation of the true risk $J(u(\bd{x} \, ; \Theta))$.

\section{Our Method}
For simplicity, we assume the boundary term in \eqref{dis_loss} has been well approximated and only consider $J_r(u(\bd{x}\,;\Theta))$, the loss inside $\Omega$ induced by the residual. Instead of solving the original problem \eqref{eq:problem} with a static pre-sampled dataset by PINNs, an incremental learning view will be introduced, with which we can gradually capture the singularities in our problem.

\subsection{Incremental learning for PINNs}\label{incremental_learning}

To begin with, we first relax the definition of loss function $J_r$ with a general background distribution $\rho(\bd{x})$ as follows:
\begin{equation}
  J_r(u(\bd{x};\Theta)) = \int r^2(\bd{x};\Theta)\rho(\bd{x})dx.
\end{equation}
This relaxation, however, does not change the minimizer $u(\bd{x};\Theta^*)$ if the minimal value is zero. We may deliberately and actively select the background density $\rho(\bd{x})$ to focus on the learning of regions with higher risk, such as the neighbor of singularity. In general, different definitions of the best $\rho$ lead to different adaptive sampling strategy.

In practice, the background distribution $\rho(\bd{x})$ and the approximation solution $u(\bd{x})$ can be solved with alternate iteration. Starting from some prior distribution $\rho_0(\bd{x})$, the DAS method aims to find the best solution with the following alternate iteration:
\begin{eqnarray}
  u^{k+1} &=& \underset{u}{\arg\min} \int r^2(u(\bd{x},\theta)) \rho^k(\bd{x}) dx, \label{alt_DAS1}  \\
  \rho^{k+1} &=& \underset{\rho(x)\geq0,\int \rho(x)dx = 1}{\arg\max}-D_{KL}(\rho(\bd{x})\|\frac{r^2(u^{k+1})}{\int r^2 dx}), \label{alt_DAS2}
\end{eqnarray}
where $D_{KL}(\rho_1\|\rho_2)$ is the Kullback-Leibler (KL) divergence between $\rho_1$ and $\rho_2$. On the other hand, the FI-PINN method proposed to add new points by importance sampling for the approximation of failure probability:
\begin{eqnarray}
  u^{k+1} &=& \underset{u}{\arg\min} \int r^2(u(\bd{x},\theta)) \rho^k(\bd{x}) dx, \label{alt_fippin1}\\
  \rho^{k+1} &=& \underset{\rho(x)\geq0,\int \rho(x)dx = 1}{\arg\min}\|\rho(\bd{x}) - I_{r(u^{k+1})>\epsilon}(\bd{x})\|, \label{alt_fippin2}
\end{eqnarray}
here $ I_{r(u^{k+1})>\epsilon}(\bd{x})$ is the indicator function over the failure region (e.g., the region where residual is larger than $\epsilon$), of which the integral is defined as the failure probability. 

During the alternate iteration, the new data can be added in the exited dataset, rather than a simple replacement. This indicates that the distribution $\rho$ can also be updated in an incremental way: 
\begin{equation}\label{up_rho_1}
  \rho^{k+1}(\bd{x}) = \alpha_{k}\rho_{old}(\bd{x}) + (1-\alpha_{k})\rho_{add}(\bd{x}).
\end{equation}
The first term $\rho_{old}\triangleq \rho_k$ in $\eqref{up_rho_1}$ stands for the data distribution used in previous training, with which we can review the learned knowledge and avoid the catastrophic forgetting phenomenon\cite{french1999catastrophic}. The second term is proposed to adaptively sample new points for a better description of the risk, with which we can localize the singularity and sample more points in high risk region. $\alpha_k$ is a hyperparameter introduced to balance the ratio of new knowledge and old knowledge.

\subsection{Active learning with risk maximization}\label{active_learning}

Instead of using the importance sampling view for the update of $\rho$ as in DAS and FI-PINN, we would formulate the adaptive sampling (updating $\rho$) as the problem of risk maximization, following the spirit of the least confident strategy in the literature of active learning \cite{settles2009active}. In this case, the learning process can be formulated as:

\begin{eqnarray}
  u^{k+1} &=& \underset{u}{\arg\min} \int r^2(u(\bd{x},\theta)) \rho^k(\bd{x}) dx,   \label{alt_gas1}\\
  \rho_{add}(\bd{x}) &=& \underset{\rho(x)\geq0,\int \rho(x)dx = 1}{\arg\max}\int  r^2(u^k(\bd{x},\theta))\rho(\bd{x})dx, \label{alt_gas2}\\
  \rho^{k+1}(\bd{x}) &=& \alpha_{k}\rho_{old}(\bd{x}) + (1-\alpha_{k})\rho_{add}(\bd{x}). \label{alt_gas3}
\end{eqnarray}

Apart from the simplicity in form, one additional advantage of this method is that, the alternate iteration in \eqref{alt_gas1}-\eqref{alt_gas3} can be regarded as an iterative optimization for a min-max problem:
\begin{eqnarray}
u^*, \rho^* &=& \arg \underset{u}{\min} \underset{\rho(x)>0,\int \rho(x)dx = 1}{\max}L(u,\rho), \\
   &\triangleq& \arg \underset{u}{\min} \underset{\rho(x)>0,\int \rho(x)dx = 1}{\max}\int r^2(u)\rho(\bd{x})dx.
\end{eqnarray}\label{min_max}

This is similar to the training process of Generative Adversarial Nets (GAN,\cite{goodfellow2020generative}). With such formulation, the effectiveness of our adaptive sampling can be studied in the view of solving min-max problem. We will leave this part as our next work. It should also be pointed out that, the alternate iteration in \eqref{alt_DAS1}\eqref{alt_DAS2} is not equal to solving the following min-max problem:
\begin{eqnarray}
u^*, \rho^* &=& \arg \underset{u}{\min} \underset{\rho}{\max}L_{DAS}(u,\rho), \\
   &\triangleq& \arg \underset{u}{\min} \underset{\rho}{\max}\int r^2(u) \rho(\bd{x})dx- D_{KL}(\rho(\bd{x})\|\frac{r^2(\bd{x};\Theta)}{\int r^2 dx}),\label{min_max_das}
\end{eqnarray}
since both the residual loss and the KL-divergence depend on $\rho$ and $u$. In fact, it is non-trivial to give an explicit form of the min-max problems corresponding to the alternate iteration method used in DAS and FI-PINN, if they exit. In the meanwhile, one could use the above \eqref{min_max_das} to build a new adaptive sampling method based on DAS, in which the new distribution are generated to maximize the risk and minimize the distance to residual at the same time.

In general, without any constraints on the density function $\rho(\bd{x})$, the risk maximization problem in \eqref{alt_gas1} leads to the following trivial solution:
\begin{equation}\label{trivial_p}
  \rho^*(\bd{x}) = \delta_{\bd{x}_0}(\bd{x})
\end{equation} 
in which $r(\bd{x},\Theta)$ achieves the maximum value at $\bd{x}=\bd{x}_0$. This can be seen from
\begin{eqnarray*}
  \int r^2(\bd{x};\Theta)\rho(\bd{x})dx &\leq& \max_{\bd{x}}  r^2(\bd{x};\Theta) \int \rho(\bd{x})dx \\
   &\leq & r^2(\bd{x}_0;\Theta).
\end{eqnarray*}
The singularity of the delta function $\delta_{\bd{x}_0}(\bd{x})$ makes it useless in practical since we only add new points at the maximal point of loss, given the current network parameter $\Theta$. This pointwise correction would make the learning of $u^*(\bd{x};\Theta)$ in PINNs rather inefficient. In the following, we would propose a Gaussian mixture (GM) distribution-based incremental learning paradigm, in which the distribution of added points is restricted to some special function space for a more efficient sampling.

\subsection{GAS: a Gaussian mixture distribution-based adaptive sampling method}\label{gas}

\begin{figure}
    \begin{center}
        \includegraphics[width=1.0\textwidth]{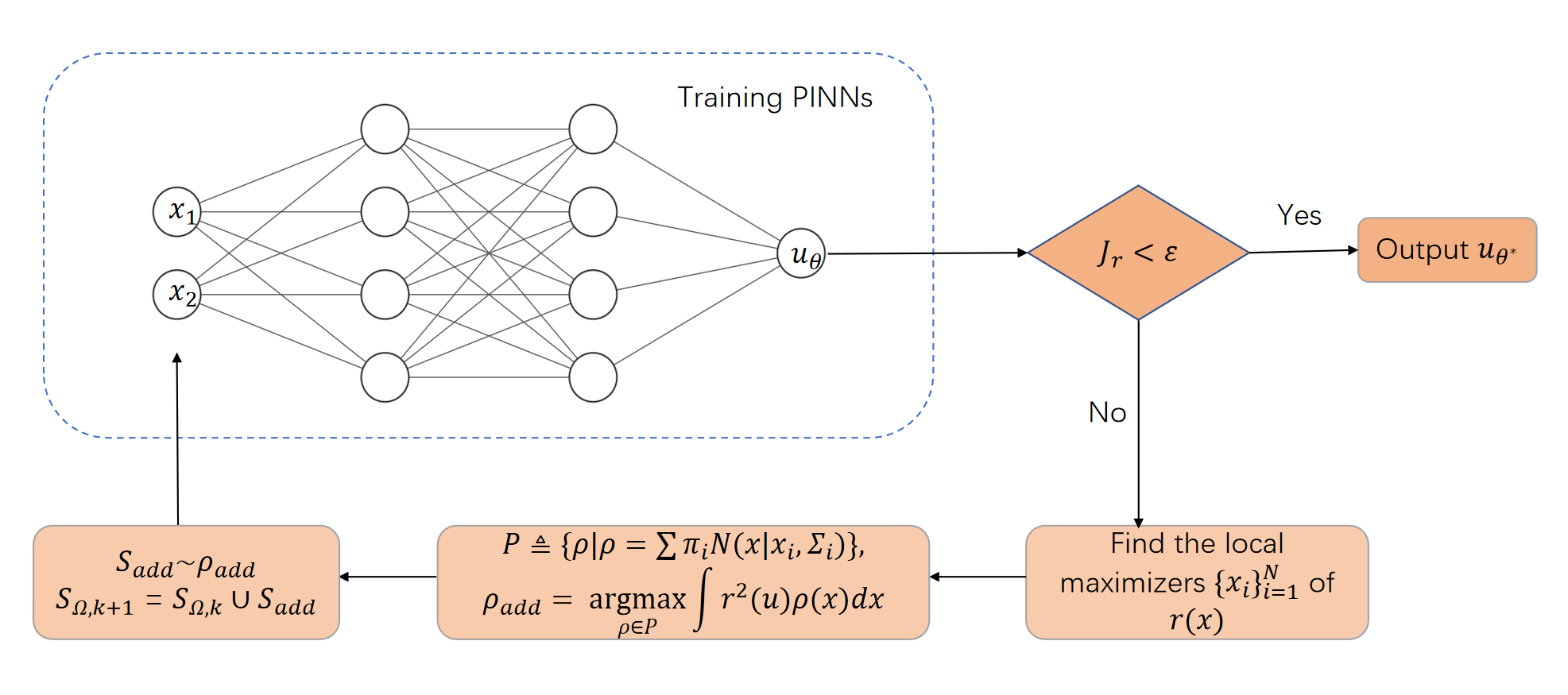}
	    \caption{The flowchart  of GAS method.}
        \label{flowchart}
    \end{center}
\end{figure}

Starting from the uniform distribution $\rho_0(\bd{x})$, we would sequentially trained PINNs and then adaptively sample and add new points according to some distribution, with which we can update the evaluation of loss and train the next round of PINNs. Since the mixture of Gaussians is a simple and widely used distribution with the ability of modeling probability densities \cite{bacharoglou2010approximation, nguyen2020approximation}, we would use it to represent the distribution of added points:
\begin{equation}\label{up_rho_2}
 \rho_{add}(\bd{x};\eta^k) \triangleq \sum_{i=1}^{N_G}\pi^k_i\mathcal{N}(\bd{x}|\mu_i^k,\Sigma_i^k),
\end{equation}
in which $\eta^k =\{\pi_i^k,\mu_i^k,\Sigma_i^k\}$ are tunable parameters and $\mathcal{N}(\bd{x}|\mu_i^k,\Sigma_i^k)$ is a Gaussian distribution with average $\mu_i^k$ and covariance $\Sigma_i^k$ . 

Following the spirit of adaptive finite element method, we use the residual function $r(\bd{x};\Theta)$ as an indicator to determine the position of refined region. Taking the one-dimensional case as an example, the means and variance appeared in Gaussian can be determined by solving the following risk maximization problem:
\begin{equation}\label{max_risk}
  \mu^*, \sigma^* = \arg\max_{\mu,\sigma}\int r^2(x;\Theta)\mathcal{N}(x|\mu,\sigma) dx
\end{equation}

To derive an explicit expression for $\mu^*$ and $\sigma^*$, we would further simplify the calculation in \eqref{max_risk} by using the idea of Laplace approximation \cite{bishop2006pattern}. Suppose $r(x;\Theta)$ has a unique local maximum point at $x = x_0$, by writing $r(x;\Theta)$ as 
\begin{equation}\label{logR}
  r(x;\Theta) = e^{-V(x)}
\end{equation} 
and taking the second order Taylor's expansion of $V(x)$ at $x_0$, we have
\begin{eqnarray}
  r(x;\Theta) &=& e^{-[V(x_0) + V^{'}(x_0) (x-x_0)+ \frac{1}{2}V^{''}(x_0)(x-x_0)^2]}\\
   &=& C\cdot e^{-\frac{1}{2}V^{''}(x_0)(x-x_0)^2} \label{lap_r}
\end{eqnarray}
Here $C$ is constant independent of $x$ and we have used the fact $V^{'}(x_0) = 0$ since $x_0$ is a maximum point. Denote $V^{''}(x_0) = \frac{1}{2a^2}$ and Substitute \eqref{lap_r} into \eqref{max_risk}, we have 
\begin{eqnarray}
    \mu^*, \sigma^* &=& \arg\max_{\mu,\sigma}\int r^2(x;\Theta)\mathcal{N}(x|\mu,\sigma) dx \\
   &=& \arg\max_{\mu,\sigma}\frac{C^2}{\sqrt{2\pi}\sigma}\int e^{-[V^{''}(x_0)(x-x_0)^2+\frac{(x-\mu)^2}{2\sigma^2}]} dx\\
   &=& \arg\max_{\mu,\sigma}\frac{1}{\sigma}\int e^{-[\frac{(x-x_0)^2}{2a^2}+\frac{(x-\mu)^2}{2\sigma^2}]} dx \\
   &=& \arg\max_{\mu,\sigma}\frac{1}{\sigma}\int e^{-[(\frac{1}{2\sigma^2}+\frac{1}{2a^2})(x - \frac{a^2\mu + \sigma^2 x_0}{a^2+\sigma^2})^2 + \frac{(\mu-x_0)^2}{2a^2+2\sigma^2}]} dx \\
   &=& \arg\max_{\mu,\sigma}\frac{e^{-\frac{(\mu-x_0)^2}{2a^2+2\sigma^2}}}{\sigma}\int e^{-[(\frac{1}{2\sigma^2}+\frac{1}{2a^2})(x - \frac{a^2\mu + \sigma^2 x_0}{a^2+\sigma^2})^2]} dx \\
   &=& \arg\max_{\mu,\sigma}\frac{1}{\sigma}\sqrt{\frac{2\sigma^2a^2}{\sigma^2+a^2}}\cdot e^{-\frac{(\mu-x_0)^2}{2a^2+2\sigma^2}} \\
   &=& \arg\max_{\mu,\sigma}\sqrt{\frac{2}{1 + (\sigma/a)^2}} \cdot e^{-\frac{(\mu-x_0)^2}{2a^2+2\sigma^2}} \label{last_mu}
\end{eqnarray}

As mentioned in the end of last subsection \ref{active_learning}, without any constraint of the covariance $\sigma$, the best proposal density $\rho_{add}$ will collapse to the delta function $\delta_{x_0}(x)$. This can be seen from the last term in \eqref{last_mu}, which can only take the maximum when $\sigma = 0$. To avoid this, we restrict the covariance to that $\sigma \geq a$. This restriction, actually, extends the search of high risk points to a bigger region rather than a single points, thus supplying us with a more efficient strategy especially when the local maximizer $x_0$ is poorly approximated. With this constraint, the maximum value in \eqref{max_risk} can obtained at 
\begin{equation}\label{opt_mu}
  \mu = x_0, \quad \sigma = a = [2V^{''}(x_0)]^{-1/2}
\end{equation}

According to \eqref{logR} and the fact that $r^{'}(x_0) = 0$, $V^{''}(x_0)$ can be calculated as 
\begin{equation}\label{V2r}
  V^{''}(x_0) = - \frac{r^{''}(x_0)}{r(x_0)}
\end{equation}

On the other hand, since a second order derivative of $r$ may require two times back propagation in the network computation, which would make the searching of $\rho_{add}$ inefficient, we would replace the $r^{''}(x_0)$ with its difference approximation:
\begin{equation}\label{diff_d2r}
  V^{''}(x_0)\approx- \frac{r^{'}(x_0+\epsilon)}{\epsilon \cdot r(x_0)}
\end{equation}

In practice, the maximum point $x_0$ of $r(x)$ can be approximated by the sample $\hat{x}_0$, which is the maximum point of $r(x)$ obtained in a finite dataset, e.g., all the sample data used in the k-round training of PINNs. In this case, we may hope the difference $|x_0-\hat{x}_0| = O(\epsilon)$, and compute the variance $\sigma$ as 
\begin{equation}\label{sigma_app}
  \sigma = [2V^{''}(x_0)]^{-1/2} = |\frac{r(\hat{x}_0)}{2}\epsilon |^{1/2} \cdot |r^{'}(\hat{x}_0)|^{-1/2}.
\end{equation}  
A formal understanding of the selection in \eqref{opt_mu}\eqref{sigma_app} is that, we hope to sample additional data around the points with highest risk, and a more rapidly decrease of the local risk indicates less points should be sampled far away from the means. 

In multi-peak and high dimensional case, by using the Laplace approximation to $V(\bd{x})$ with a diagonal simplification in the calculation of covariance matrix,  we may similarly generate the proposed density $\rho_{add}$ as mixture of Gaussians, with the means and variances defined as follows:
\begin{eqnarray}
  \mu_i^k &=& \underset{x\in U\cap S_{\Omega,k}}{\arg\max} r(\bd{x};\Theta),\label{mean}\\
  \Sigma_i^k &=& \lambda [diag(\nabla_xr(\bd{x};\Theta)|_{x = u_i^k})]^{-1}. \label{cov}
\end{eqnarray}
Here $S_{\Omega,k}$ is the dataset used in the k-round training of PINN, $U$ stands for some open set, thus $\mu_i^k$ is a local maximizer of the residual function. $\lambda$ is a hyperparameter and $diag(\alpha)$ refers to diagonal matrix with the diagonal element vector being $\alpha$. 

In practice, one can use the top $N_G$ points with largest residual, instead of using the $N_G$ local maximum point of $r(x)$ to define the means. This compromise, as we will see, has little influence on the effectiveness of our method, if only that the number of singularities is not too much and $N_G$ is not too small. With the new distribution $\rho_{k+1}= \alpha_k \rho_k + (1-\alpha_k)\rho_{add}$, the evaluation of $J_r$ can be updated as
\begin{equation}\label{up_loss}
  J_r^{k+1}(u(\bd{x};\Theta) = \int r^2(\bd{x}; \Theta) \rho_{k+1}(\bd{x})dx.
\end{equation}
Then we may repeat the training process of PINNs by minimizing $J_r^{k+1}$. The flowchart of our method is shown in Figure \ref{flowchart}.

A detail description of our idea is presented in Algorithm \ref{alg:GAS-algorithm}, in which we divide the GAS methods into GAS-T (define the means by top $N_G$ points) and GAS-L (define the average by local maximizers). As shown in the next section, this simple adaptive sampling strategy supply us with an effective and easy-to-implement method for solving elliptic equations with singularity, in both low dimension and high dimension case.

\begin{algorithm}
   \caption{GAS}
   \label{alg:GAS-algorithm}
\begin{algorithmic}
   \STATE {\bfseries Input:} $u(\bd{x}\,;\Theta^{(0)})$, maximum epoch number for PINNs $N_p$, number of times for adaptive sampling $N_a$, number of Gaussians in the mixture distribution $N_G$, batch size $m$, positive parameters $\lambda$, initial training set $S_{\Omega,0}=\{\bd{x}_{\Omega,0}^{(i)}\}_{i=1}^{N_r}$ and $S_{\partial \Omega,0}=\{\bd{x}_{\partial \Omega,0}^{(i)}\}_{i=1}^{N_b}$, validated set $S_{valid} = \{\bd{y}^{(i)}\}_{i=1}^{N_t}$
   \STATE {\bfseries Output:} $u(\bd{x}\,;\Theta^*_N)$
   \WHILE{$k\leq N_a-1$ and $J_r\leq \epsilon$}
   \FOR{$i=0$ {\bfseries to} $N_p - 1$}
   \FOR{$j$ {\bfseries steps}}
   \STATE Sample $m$ samples from $S_{\Omega, k}$.
   \STATE Sample $m$ samples from $S_{\partial \Omega, k}$.
   \STATE Update $u(\bd{x}\,;\Theta)$ by SGD of $J_N(u(\bd{x}\,;\Theta))$.
   \ENDFOR
   \ENDFOR
   \STATE Calculate all residuals in $\{\bd{y}^{(i)}\}_{i=1}^{N_t}$ by (\ref{dis_loss}).
   \STATE Sort the validated points according to the residuals in a descending order, keep the top $N_G$ terms $\tilde{y}^{(1)}, \tilde{y}^{(2)}, \cdots, \tilde{y}^{(N_G)}$ (GAS-T) or find the $N_G$ local maximizers $\tilde{y}^{(1)}, \tilde{y}^{(2)}, \cdots, \tilde{y}^{(N_G)}$ (GAS-L).
   \FOR{$t=0$ {\bfseries to} $N_{G} - 1$}
   \STATE Construct a Gaussian distribution $\mathcal{N}^{(t)}(\mu, \Sigma)$ by the coordinates of $\tilde{y}^{(t)}$ as the mean $\mu$ and the gradient of residual to its coordinates as the reciprocal of the covariance $\Sigma$.
   \STATE Generate one or more points, denoted as the set $S_g^{(t)}$, which are sampled from $\mathcal{N}^{(t)}(\mu, \Sigma)$.
   \ENDFOR
   \STATE Add all the points so that $S_{\Omega, k+1} = S_{\Omega, k} \bigcup S_g^{(0)} \bigcup S_g^{(1)} \cdots \bigcup S_g^{(N_{G} - 1)}$
   \STATE Add points in boundary from uniform distribution so that the ratio of the points in $S_{\Omega, k}$ to the points in $S_{\partial \Omega, k}$ is a constant.
   \ENDWHILE
\end{algorithmic}
\end{algorithm}

\section{Numerical experiments}
\subsection{One-peak problem}
\begin{example}\label{Exam_1}
As a benchmark test, we first consider the following elliptic problem with the GAS-T method.
\begin{equation}\label{eq:problem_1}
  \left\{
  \begin{array}{ll}
  -\Delta u(x_1, x_2)&=s(x_1, x_2), \, \text{in} \,\, \Omega, \\[1.5ex]
  u(x_1, x_2)&=g(x_1, x_2), \, \text{on} \,\, \partial \Omega, \\[1.5ex]
  \end{array}
\right.
\end{equation}
where $\Omega = [-1,1]^2$. The reference solution is chosen as follows:
\begin{equation*}
  u(x_1, x_2) = \exp(-1000[(x_1-r_c)^2 + (x_2-r_c)^2]),
\end{equation*}
which has a peak at the point $(r_c, r_c)$ and decreases rapidly away from $(r_c, r_c)$. 
\end{example}

\begin{table}
\centering
\caption{MSE for different methods and point sets in one-peak problem.}
\label{MSE-one-peak}
\setlength{\tabcolsep}{5.5mm}{
\begin{tabular}{|l|c|c|c|c|}
\hline
\diagbox{Strategy}{MSE}{$|S_{\Omega}|$} & $2000$ & $3000$ & $4000$ & $5000$ \\
\hline
Uniform & 7.5E-03  & 7E-03  & 3E-03  & 1E-03   \\
\hline
DAS     & 4E-03  & 8E-04  & 7.5E-04  & 4E-04   \\
\hline
GAS     & 1.9E-04  & 1.8E-04  & 3.4E-05  & 1.0E-05   \\
\hline
\end{tabular}}
\end{table}

Due to the highly singularity at the peak point, a uniform distribution-based loss would make the PINNs rather inefficient.  In this example, we choose a six-layer fully connected neural network $u(\bd{x}\,; \Theta)$ with $32$ neurons to approximate the solution, and uniformly sample $N_r = 500$ points in $\Omega$ and $N_b = 200$ points on the boundary. During the incremental learning, we train $N_p = 3000$ epochs for PINNs after each adaptive sampling and the number of adaptive sampling times is set as $N_a = 10$. The batch size $m$ is chosen as $500$ for the points in $S_{\Omega}$ and $200$ for boundary samples. 

As for parameters in GAS, we set $N_{G} = 20$ to construct the Gaussian mixture model and sample $25$ points from each Gaussian, thus a total of $500$ points would be added into $S_{\Omega}$ during each adaptive sampling. To maintain the ratio of interior and boundary points, we add $200$ points additionally into $S_{\partial \Omega}$, which are simply sampled by uniform distribution.

\begin{figure}[htb]
    \begin{center}
        \includegraphics[width=0.9\textwidth]{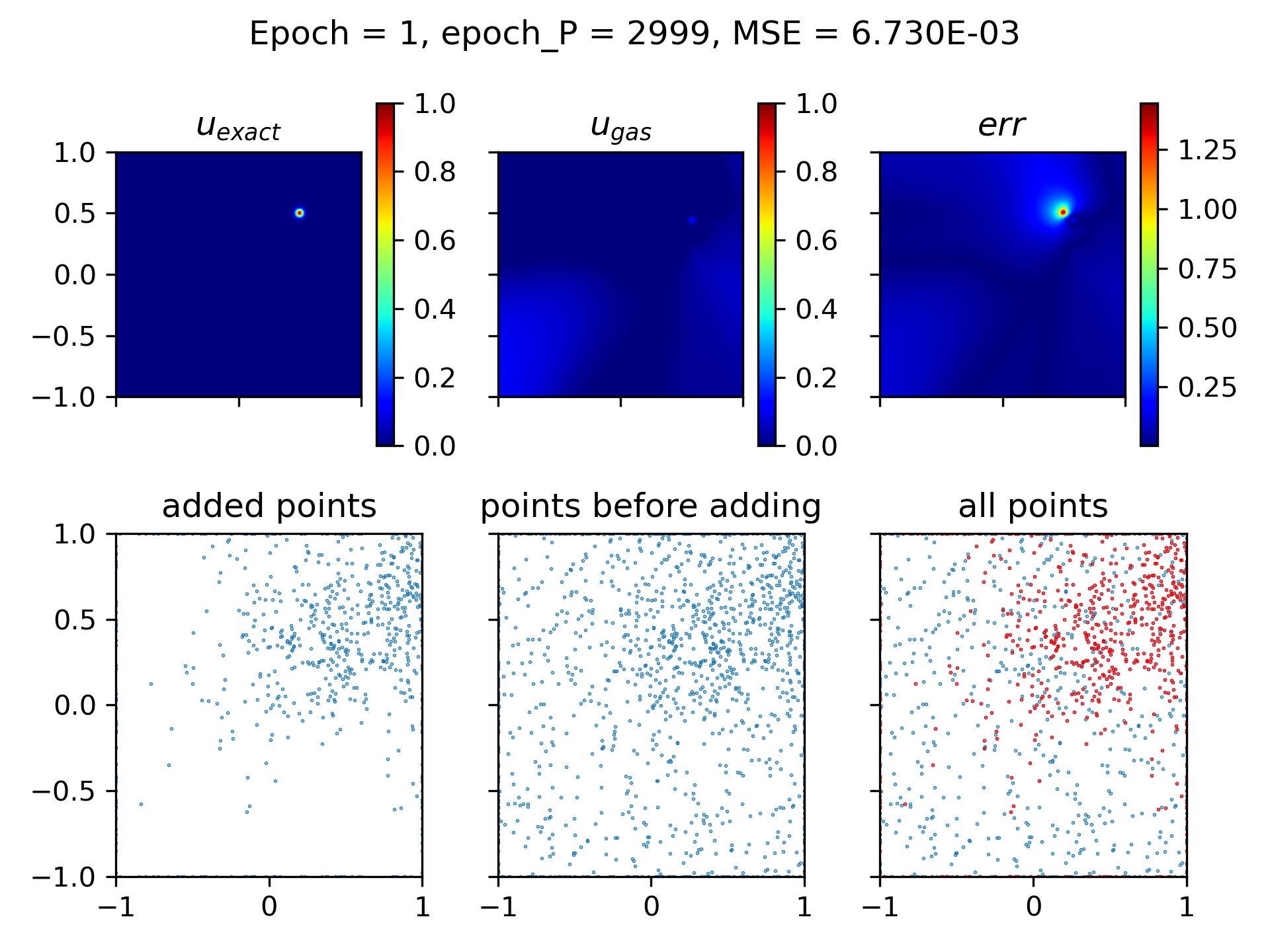}
	    \caption{PINNs with 1st adaptive sampling points in one peak problem and $|S_{\Omega}| = 1000$.}
        \label{GAS-0-2999}
    \end{center}
\end{figure}

\begin{figure}[htb]
    \begin{center}
        \includegraphics[width=0.9\textwidth]{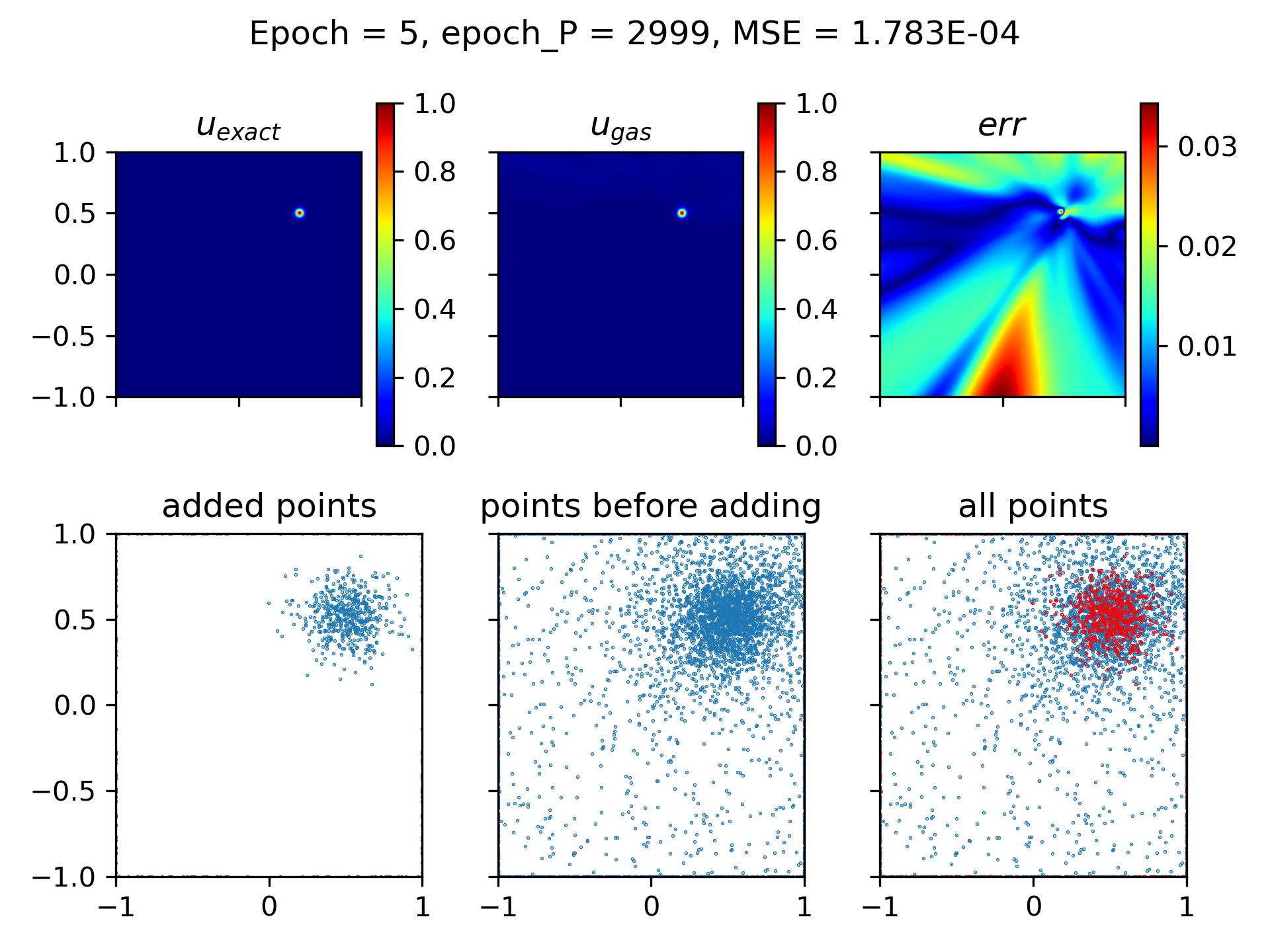}
	    \caption{PINNs with 5th adaptive sampling points in one peak problem and $|S_{\Omega}| = 3000$.}
        \label{GAS-5-2999}
    \end{center}
\end{figure}

\begin{figure}[htb]
    \begin{center}
        \includegraphics[width=0.9\textwidth]{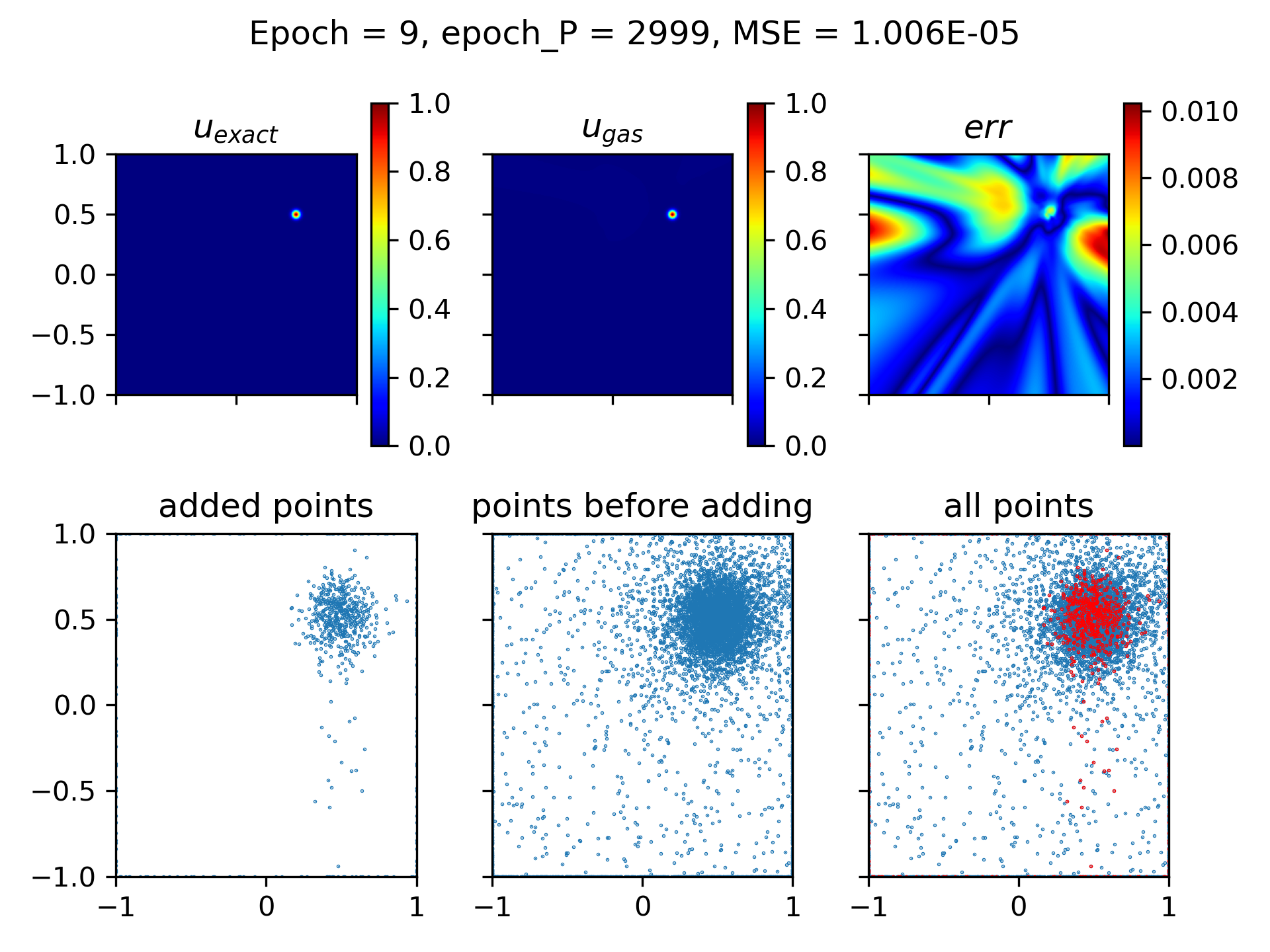}
	    \caption{PINNs with 9th adaptive sampling points in one peak problem and $|S_{\Omega}| = 5000$.}
        \label{GAS-9-2999}
    \end{center}
\end{figure}

\begin{figure}
    \begin{center}
        \includegraphics[width=0.9\textwidth]{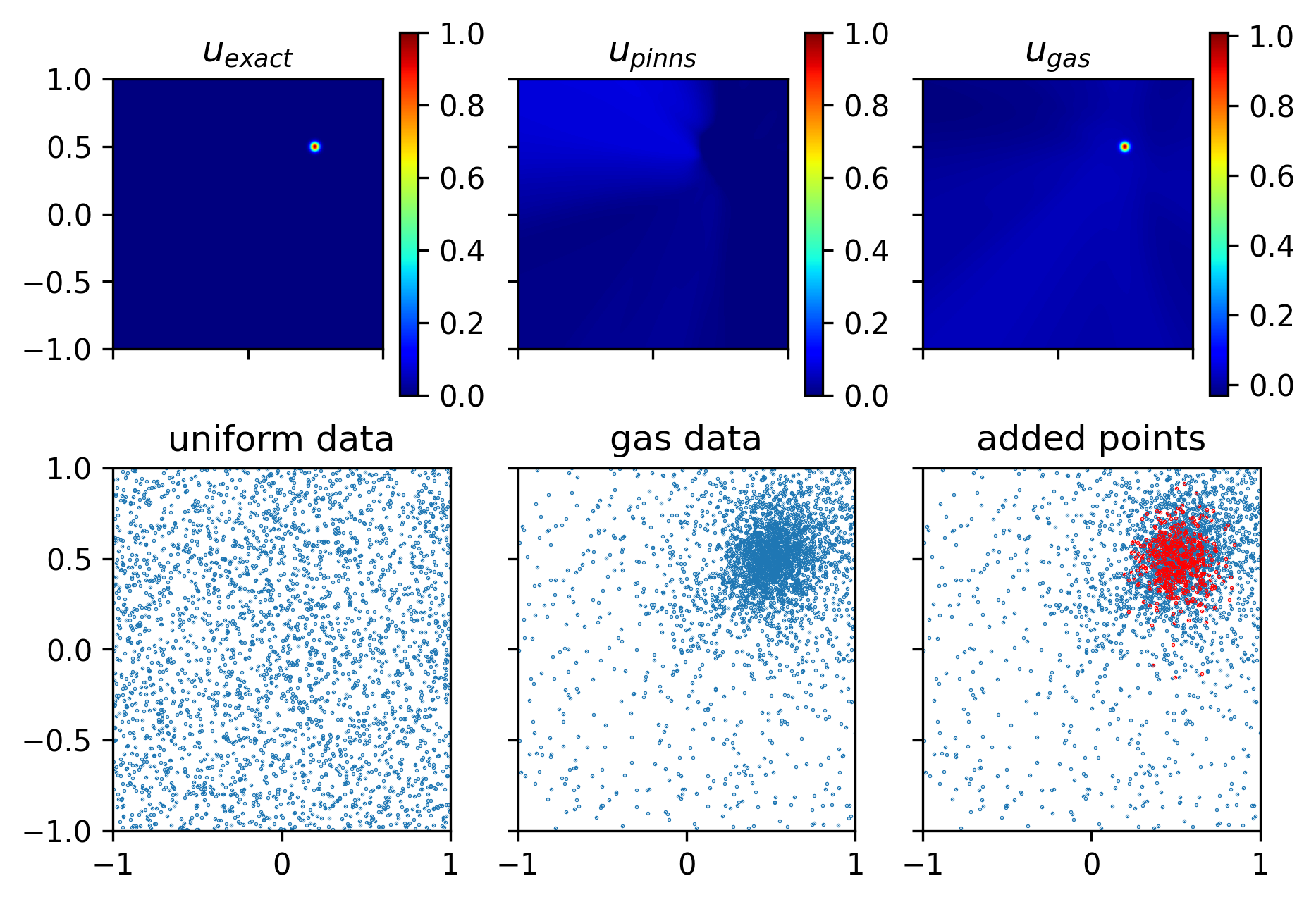}
	    \caption{Solutions and $3000$ points for uniform and GAS}
        \label{compare_uniform_2960}
    \end{center}
\end{figure}

Figure \ref{GAS-0-2999}, \ref{GAS-5-2999}, \ref{GAS-9-2999} shows the numerical result of PINNs after 1 time, 5 times and 9 times of adaptive sampling. According to these, we can see that the magnitude of error becomes smaller with the times of adaptive sampling increasing. The solution by GAS converges almost to the exact solution after 3000 samples have been used, while the PINNs method with a uniform sampling strategy fails to deal with the point singularity (see Figure \ref{compare_uniform_2960}). Since the solution owns one peak at $(0.5,0.5)$, the aggregated additional samples are mainly located around this singular point where the residual is much bigger, which is exactly we expect. A further comparison of our method with the results by using uniform sampling and DAS method is shown in Table \ref{MSE-one-peak}, which reveals that our method can achieve a uniformly better mean-square error (MSE), given a fixed sampling size of $S_{\Omega}$. Specially, the MSE by GAS may achieve $O(10^{-5}$), which is better than the $O(10^{-4})$ by DAS and $O(10^{-3})$ with uniform distribution.

\subsection{two-peaks problem}
\begin{example}\label{Exam_2}
Next we consider the following elliptic equation with two peaks:
\begin{equation}\label{eq:problem_2}
  \left\{
  \begin{array}{ll}
  -\nabla\cdot[u(x_1,x_2)\nabla(x_1^2 + x_2^2)]+\Delta u(x_1, x_2) = s(x_1, x_2), \, \text{in} \,\, \Omega, \\[1.5ex]
  u(x_1, x_2) = g(x_1, x_2), \, \text{on} \,\, \partial \Omega, \\[1.5ex]
  \end{array}
\right.
\end{equation}
where $\Omega = [-1,1]^2$. The reference solution is set as
\begin{equation*}
\begin{aligned}
  u(x_1, x_2) &= \text{exp}(-1000[(x_1-0.5)^2 + (x_2-0.5)^2]) \\
  &+ \text{exp}(-1000[(x_1+0.5)^2 + (x_2+0.5)^2]) ,
\end{aligned}
\end{equation*}
which has two peaks at the points $(0.5, 0.5)$ and $(-0.5, -0.5)$.
\end{example}

In this example, we also set the  initial size of dataset as $N_r = 500$ and $N_b = 200$, which are sampled from uniform distribution. The other parameters are selected as $N_p = 5000$, $N_a = 20$ and the batch size $m$ for points in $S_{\Omega}$ and $S_{\partial \Omega}$ are set as 500 and 200. The hyperparameters for Gaussian mixture distributions are kept as in previous example. Considering that there are two singularities in the reference solution, we would apply the GAS-L method for the adaptive sampling.

\begin{figure}[htb]
    \begin{center}
        \includegraphics[width=0.9\textwidth]{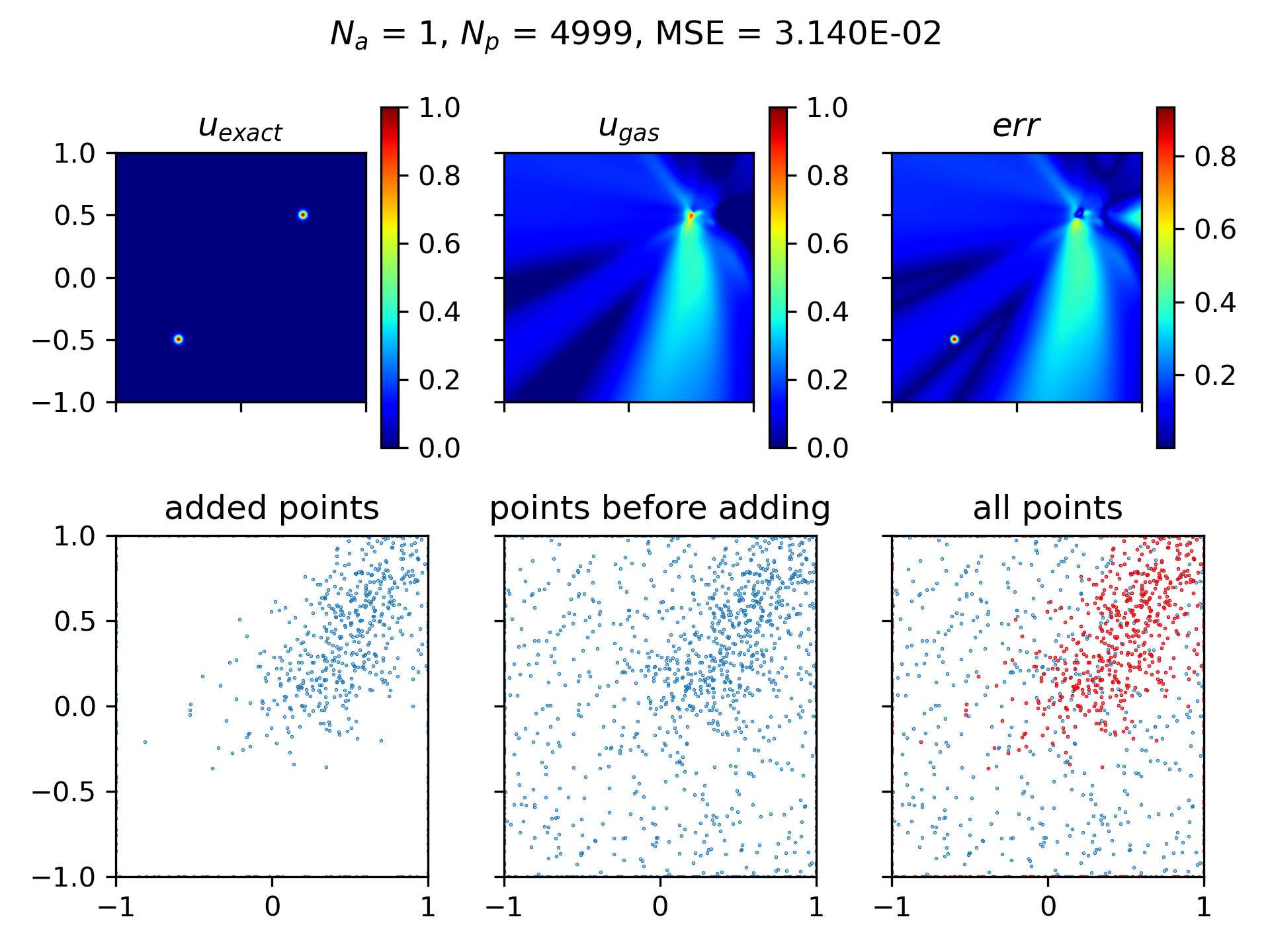}
	    \caption{PINNs with 1st adaptive sampling points in two peaks problem and $|S_{\Omega}| = 1000$.}
        \label{GAS-two-1-4999}
    \end{center}
\end{figure}

\begin{figure}[htb]
    \begin{center}
        \includegraphics[width=0.9\textwidth]{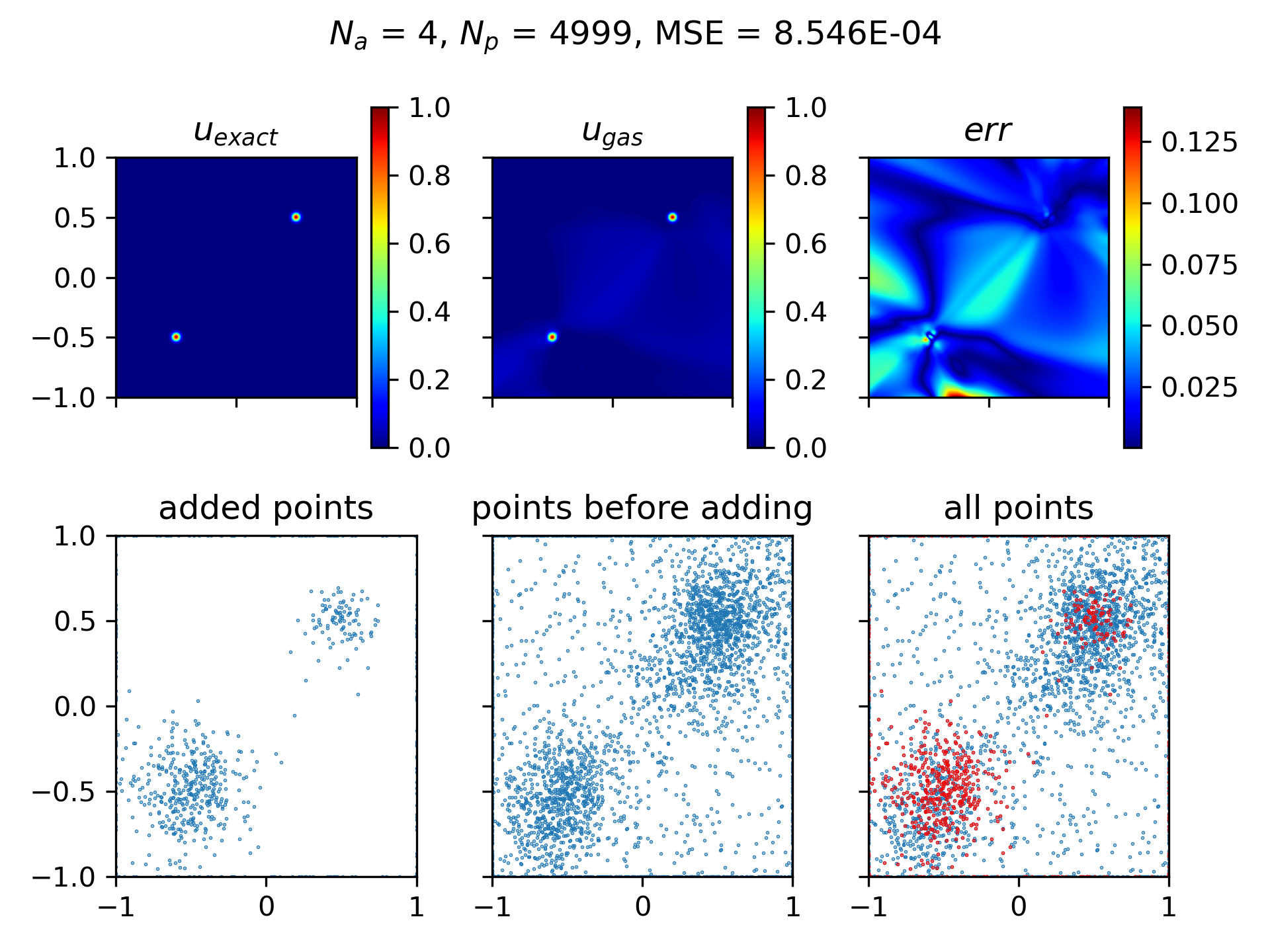}
	    \caption{PINNs with 4th adaptive sampling points in two peaks problem and $|S_{\Omega}| = 2500$.}
        \label{GAS-two-4-4999}
    \end{center}
\end{figure}

\begin{figure}[htb]
    \begin{center}
        \includegraphics[width=0.9\textwidth]{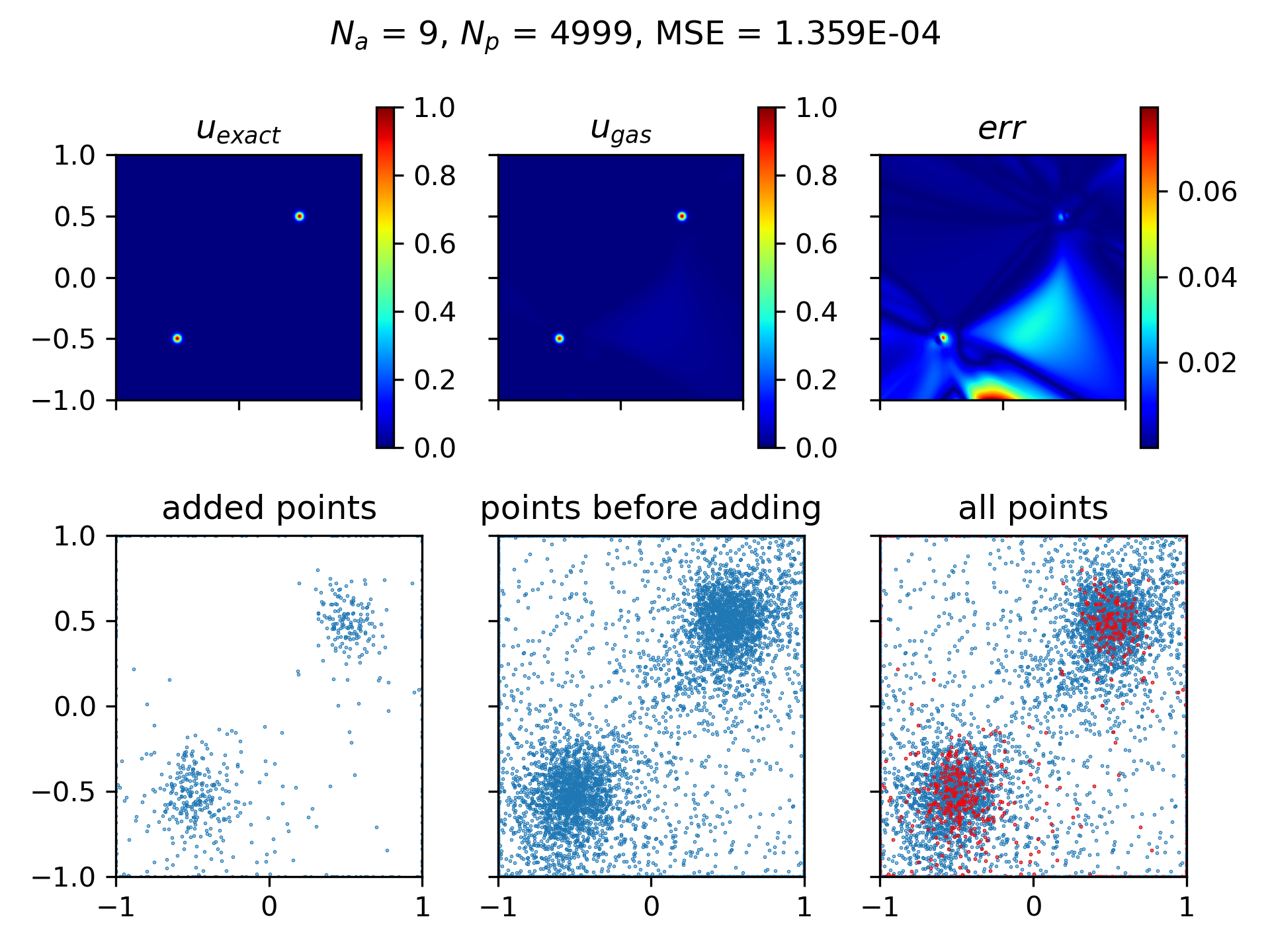}
	    \caption{PINNs with 9th adaptive sampling points in two peaks problem and $|S_{\Omega}| = 5000$.}
        \label{GAS-two-9-4999}
    \end{center}
\end{figure}

\begin{figure}[htb]
    \begin{center}
        \includegraphics[width=0.9\textwidth]{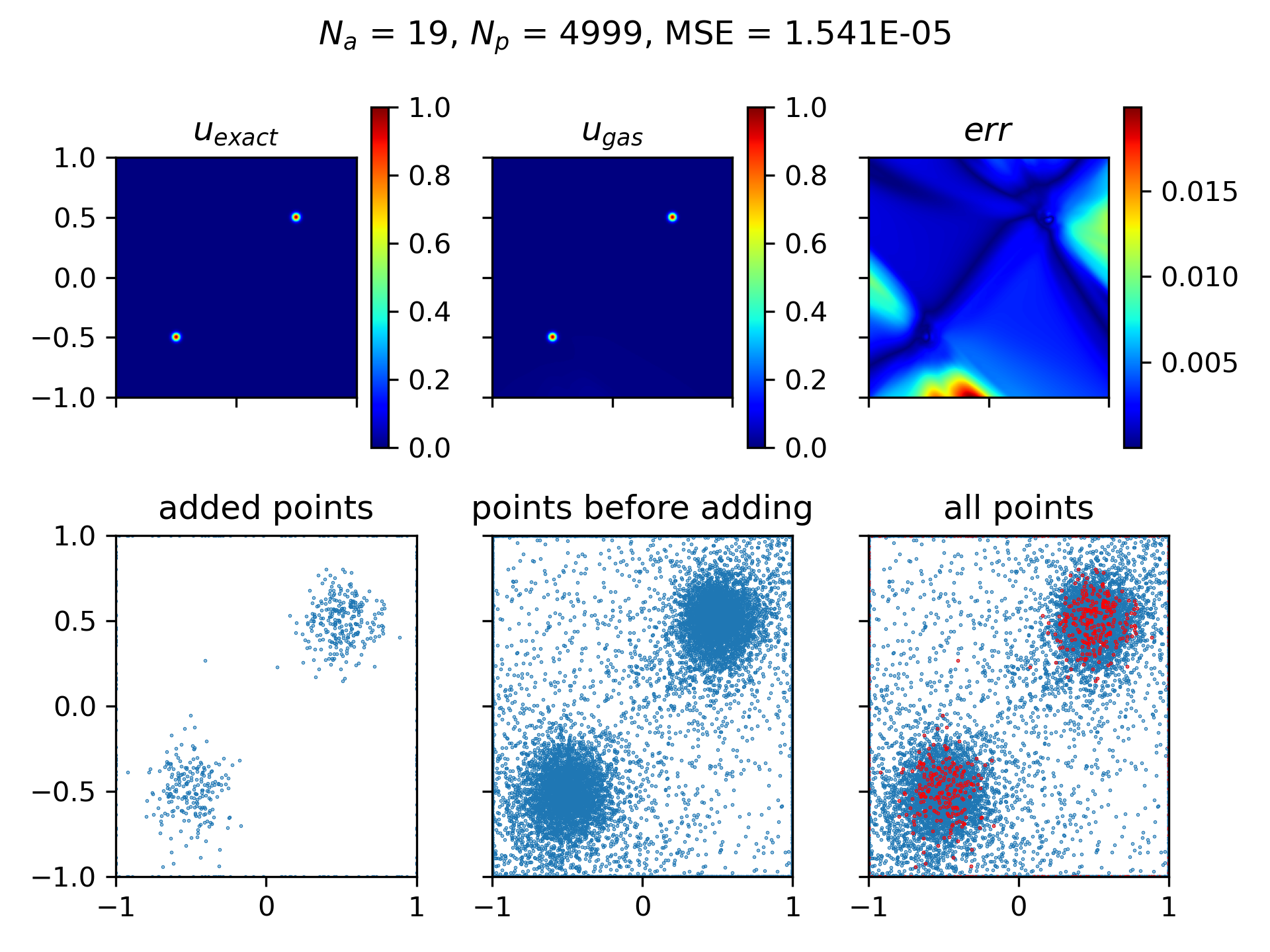}
	    \caption{PINNs with 19th adaptive sampling points in two peaks problem and $|S_{\Omega}| = 10000$.}
        \label{GAS-two-19-4999}
    \end{center}
\end{figure}

As shown in Figure \ref{GAS-two-1-4999}, \ref{GAS-two-4-4999}, \ref{GAS-two-9-4999}, \ref{GAS-two-19-4999}, the GAS method does help us to find the singularity region during the training of PINNs, while a uniform distribution-based loss can hardly be convergent. This can be understood that with more points around the two peaks in solution are sampled and added into the training dataset (see the second row of each figure), the finetuned loss functional may better represent the real risk
of the approximation by PINNs. 

In Table \ref{MSE-two-peak}, we also present the MSE along with different training size by GAS-L, DAS, uniform sampling strategy and GAS-T as well. As before, the GAS method always achieves best accuracy among these methods, and the error of GAS is one order smaller than DAS. Furthermore, since the there are two singularities in solution, the GAS-L method converges faster than GAS-T. This reflects the fact GAS-L can better describe the risk with multi singularities. In fact, during the sampling by GAS-T, the additional data is added alternately between different peaks in the early stage of learning, which leads to a slower decrease in the loss. This phenomenon, however, does not break the convergence of GAS with the adaptive sampling continuing.

\begin{table}
\centering
\caption{MSE for different methods and point sets in two-peak problem.}
\label{MSE-two-peak}
\setlength{\tabcolsep}{5.5mm}{
\begin{tabular}{|l|c|c|c|c|}
\hline
\diagbox{Strategy}{MSE}{$|S_{\Omega}|$} & $2500$ & $5000$ & $7500$ & $10000$ \\
\hline
Uniform & 7.5E-02  & 1.7E-01  & 8E-02  & 3E-02   \\
\hline
DAS     & 2E-02  & 7E-03  & 4.5E-03  & 1.5E-03   \\
\hline
GAS-T     & 6.5E-03  & 5.6E-04  & 1.7E-04  & 4.1E-05   \\
\hline
GAS-L     & 8.5E-04  & 1.4E-04  & 3.8E-05  & 1.5E-05   \\
\hline
\end{tabular}}
\end{table}

\subsection{nine-peaks problem}
\begin{example}\label{Exam_9}
To better illustrate the potential of GAS-L in dealing with multi-singularity problem, we would consider the same equation as in previous example, while the exact solution is replaced with
\begin{equation*}
  u(x_1, x_2) = \sum_{i=0}^{8}\text{exp}(-1000[(x_1-a_i)^2 + (x_2-b_i)^2]) 
\end{equation*}
in which $(a_i,b_i) = (-0.5,-0.5) + (\frac{mod(i,3)}{2},0) + (0,\frac{[i/3]}{2}), i\in\{0,...,8\}$ are nine peaks distributed in $[-1,1]^2$.
\end{example}

\begin{figure}[htb]
    \begin{center}
        \includegraphics[width=0.9\textwidth]{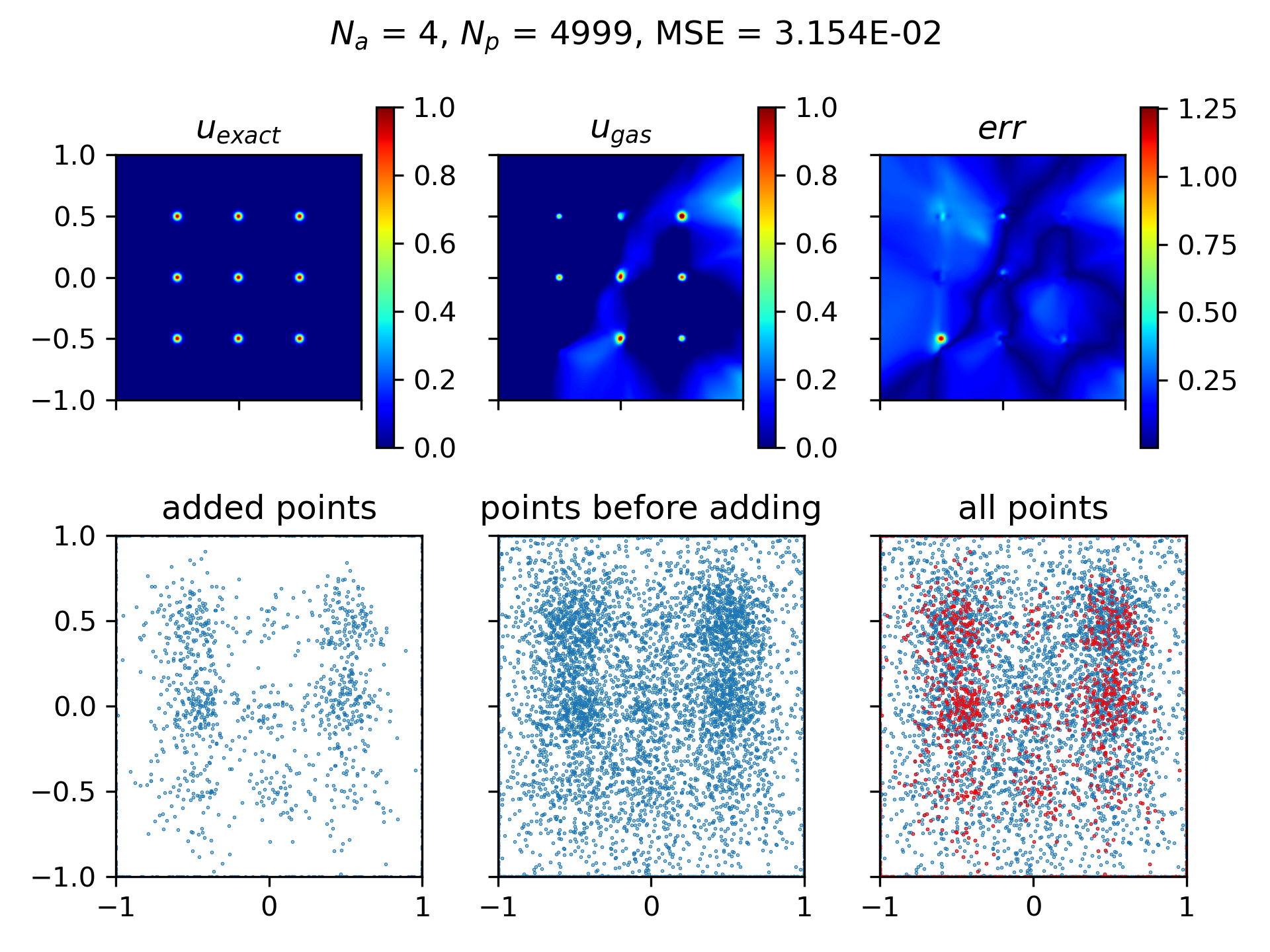}
	    \caption{PINNs with 4th adaptive sampling points in nine peaks problem and $|S_{\Omega}| = 5000$.}
        \label{GAS-nine-4}
    \end{center}
\end{figure}

\begin{figure}[htb]
    \begin{center}
        \includegraphics[width=0.9\textwidth]{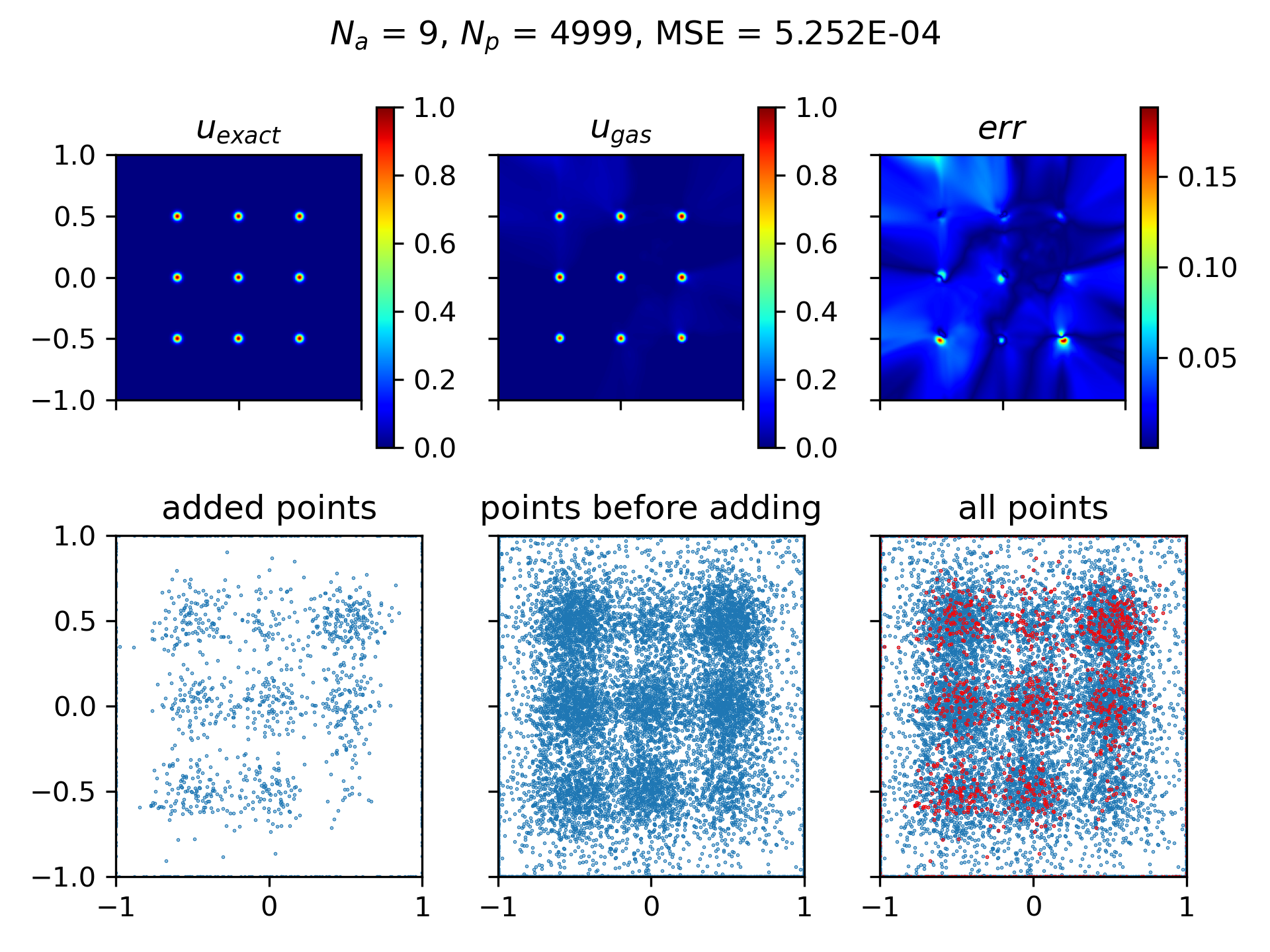}
	    \caption{PINNs with 9th adaptive sampling points in nine peaks problem and $|S_{\Omega}| = 10000$.}
        \label{GAS-nine-9}
    \end{center}
\end{figure}

\begin{figure}[htb]
    \begin{center}
        \includegraphics[width=0.9\textwidth]{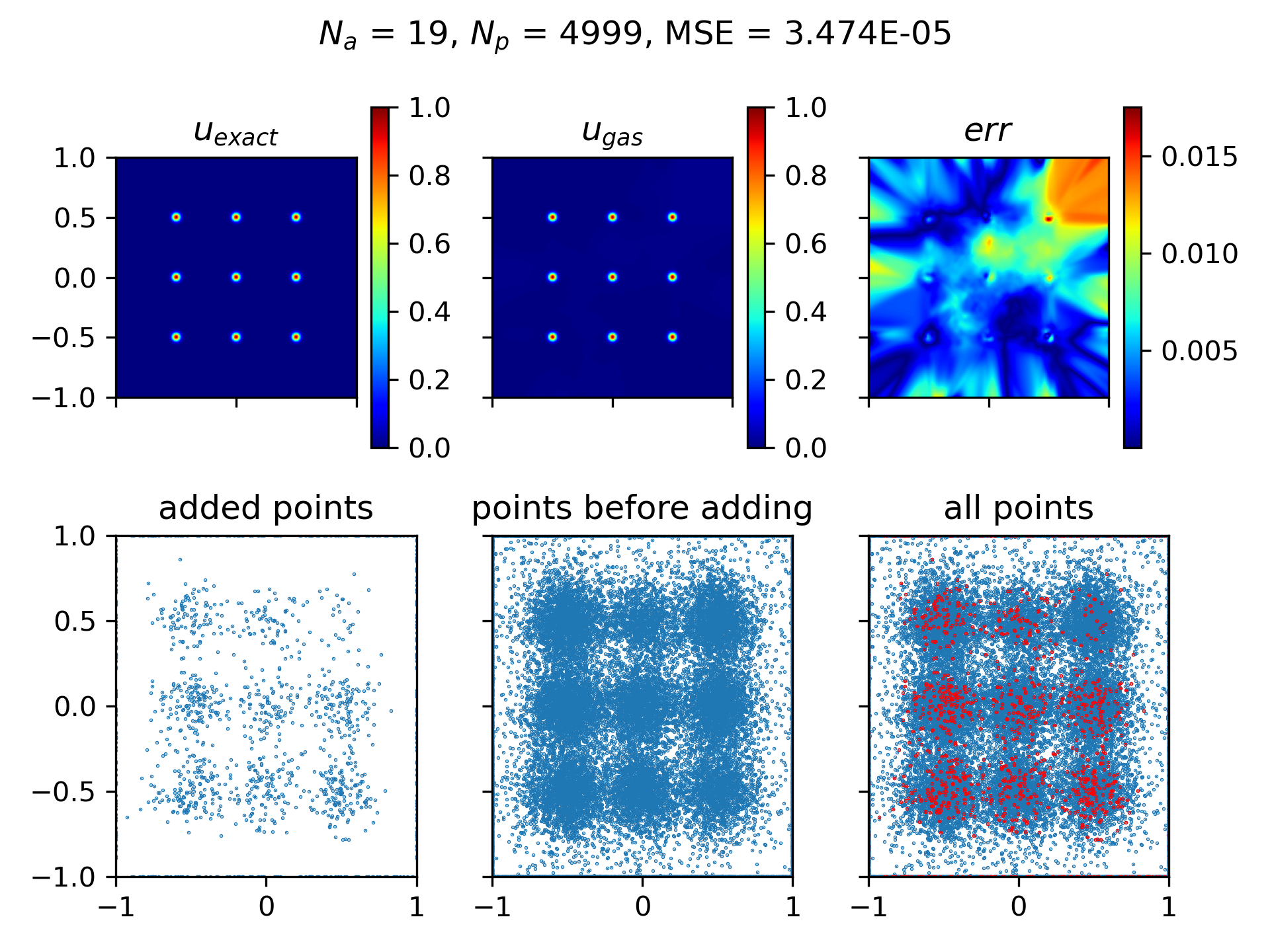}
	    \caption{PINNs with 19th adaptive sampling points in nine peaks problem and $|S_{\Omega}| = 20000$.}
        \label{GAS-nine-19}
    \end{center}
\end{figure}

As shown by Figure \ref{GAS-nine-4}, \ref{GAS-nine-9}, \ref{GAS-nine-19}, the GAS-L method can capture multi singularities at same time in adaptive sampling. After only four times incremental learning, eight peaks have been well localized by PINNs and the MSE may converges to $3.5\times 10^{-5}$ after 19 times adaptive sampling.

\subsection{High-dimensional linear problem}
\begin{example}\label{Exam_3}
As the last example, we would validate the efficiency of GAS in high dimensional case by studying the following $d$-dimensional problem:
\begin{equation}\label{eq:problem_3}
  -\Delta u(\bd{x})=s(\bd{x}), \, \bd{x} \in \Omega.
\end{equation}
where $\Omega = [-1,1]^d$. We use the following reference solution
\begin{equation*}
   u(\bd{x})=e^{-10\nm{\bd{x}}{2}^2}.
\end{equation*}
with Dirichlet boundary condition on $\partial \Omega$. 
\end{example}

In this example, we focus on the case with $d=10$. A six-layer fully connected neural network $u(\bd{x}\,; \Theta)$ with $64$ neurons would be used to approximate the solution, and the size of initial dataset are set as $N_r = 10000$ and $N_b = 10000$, which are sampled from uniform distribution. We let $N_p = 3000$, $N_a = 20$ and the batch size $m = 5000$ for the points in $S_{\Omega}$ and $S_{\partial \Omega}$. To measure the quality of approximation, a tensor grid with $n_t^d$ points are generated around the origin(in $[-0.1, 0.1]^d$) where $n_t=3$ is the number of nodes in each dimension. With these grid points, we can calculate the following relative error by using numerical integration:
\begin{equation*}
  \text{Relative error} = \frac{\nm{u_{\Theta} - u_{exact}}{2}}{\nm{u_{exact}}{2}}.
\end{equation*}
Here $u_{\Theta}$ and $u_{exact}$ denote the discrete vector of the numerical solution and exact solution. 

\begin{table}
\centering
\caption{errors for different methods in ten-dimensional linear test problem.}
\label{ND-compare}
\setlength{\tabcolsep}{1.5mm}{
\begin{tabular}{|l|c|c|r|}
\hline
\diagbox{FNS}{(error, ANS)}{strategy} & DAS & GAS \\
\hline
$5\times10^4$ & (0.030, $1.5\times10^5$) & (0.014, $1.5\times10^5$)   \\
\hline
$1\times10^5$ & (0.028, $3\times10^5$) & (0.005, $5.5\times10^5$)   \\
\hline
$2\times10^5$ & (0.008, $6\times10^5$) & (0.001, $2.1\times10^6$)   \\
\hline
\end{tabular}}
\end{table}

Since there is only one peak in our solution, we would adopt the GAS-T method for simplicity. In order to be comparable with the dataset used in \cite{tang2021das}, the number of Gaussians in the mixture model is set as $N_{G} = 40$ and we sample $250$ points from each Gaussian distribution, which supply us with $10000$ points at each time of adaptive sampling. For a better description of the size of data and cost of training, we denote the final number of different samples used by GAS as FNS , and the accumulated number of samples involved in the incremental learning as ANS. With $N_a$ times adaptive sampling and $|S_{\Omega}|$ samples added in each time, the ANS and FNS can be calculated as
\begin{eqnarray*}
  FNS &=& N_a \times |S_{\Omega}| \\
  ANS &=& \frac{FNS}{N_a} \sum_{i=1}^{N_a} i,
\end{eqnarray*}
and the total computational cost of GAS can be defined as $ANS\times N_p$.

Since the DAS-G sampling strategy takes less time and achieves smaller errors than RAR, Uniform and DAS-R (see Table 1 in \cite{tang2021das}), we would only compare the error and cost between GAS and DAS-G. As shown in Table \ref{ND-compare}, our method can achieve a smaller error even with less FNS and ANS, e.g., the error of GAS trained with $5.5\times 10^5$ points sampled from $1\times 10^5$ different samples is $5\times 10^{-1}$, which is smaller than $8\times 10^{-3}$, the error of DAS with $6\times 10^5$ points sampled from $2\times 10^5$ different samples. In general, our method can achieve better accuracy by using less data and computational cost, without mention to the fact that an additional KRnet needed in DAS may further increase the training time.

\begin{figure}[htb]
    \begin{center}
        \includegraphics[width=0.7\textwidth]{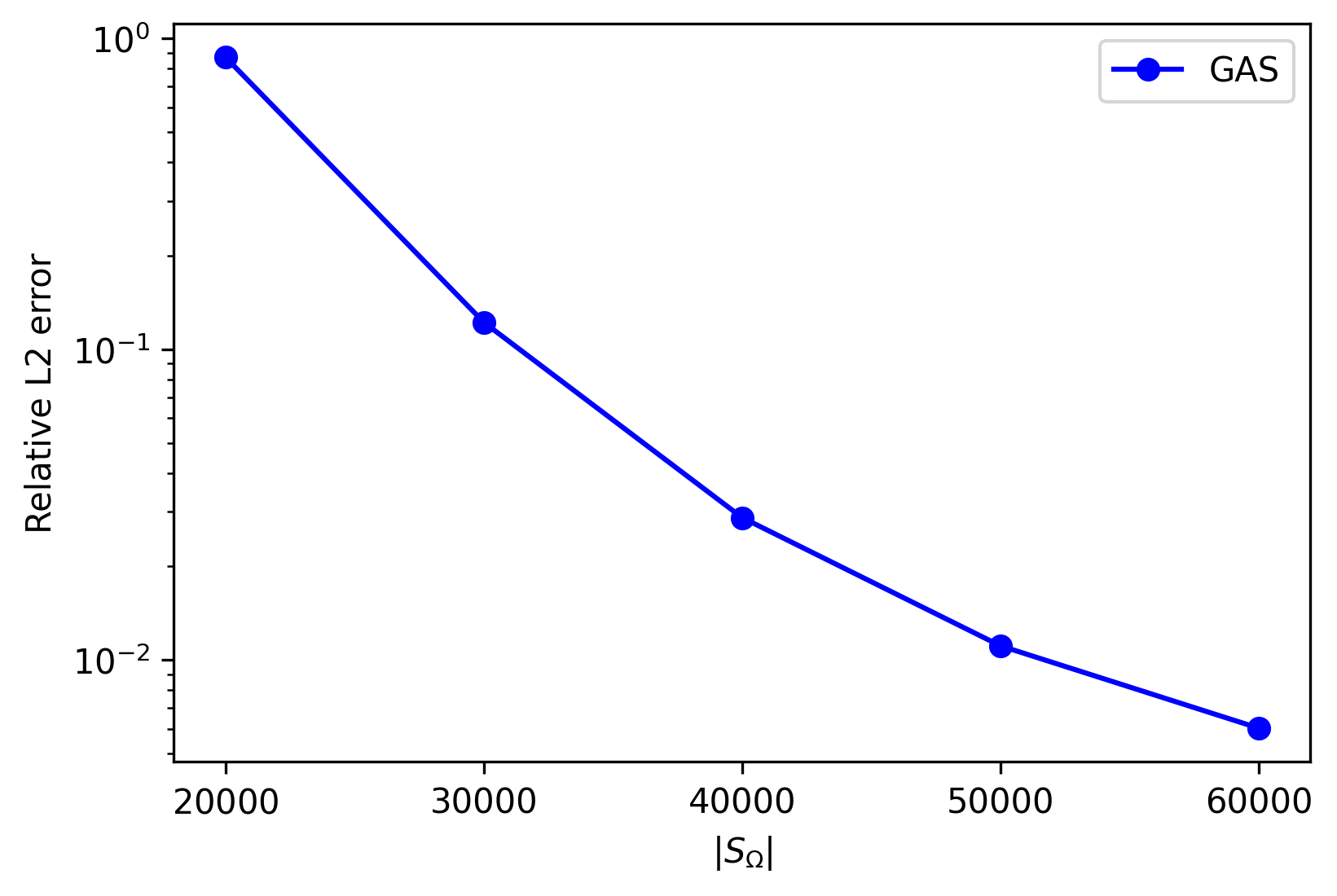}
	    \caption{Relative L2 errors for ten dimension.}
        \label{compare_FIPINNs}
    \end{center}
\end{figure}

To further demonstrate the efficiency of GAS for high dimensional problems, we also plot the relative $L_2$ error of GAS with increasing dataset, and compare it with the FIPINNs method. As shown in Figure \ref{compare_FIPINNs} and the Figure 12 in \cite{gao2022failure}, with the $|S_{\Omega}|$ increasing, the error by GAS decrease faster than that by FIPINNs, which indicates that our method may better alleviates the curse of dimensionality.

\begin{figure}[htbp]
\centering  

\subfigure[]{
\includegraphics[scale=0.46]{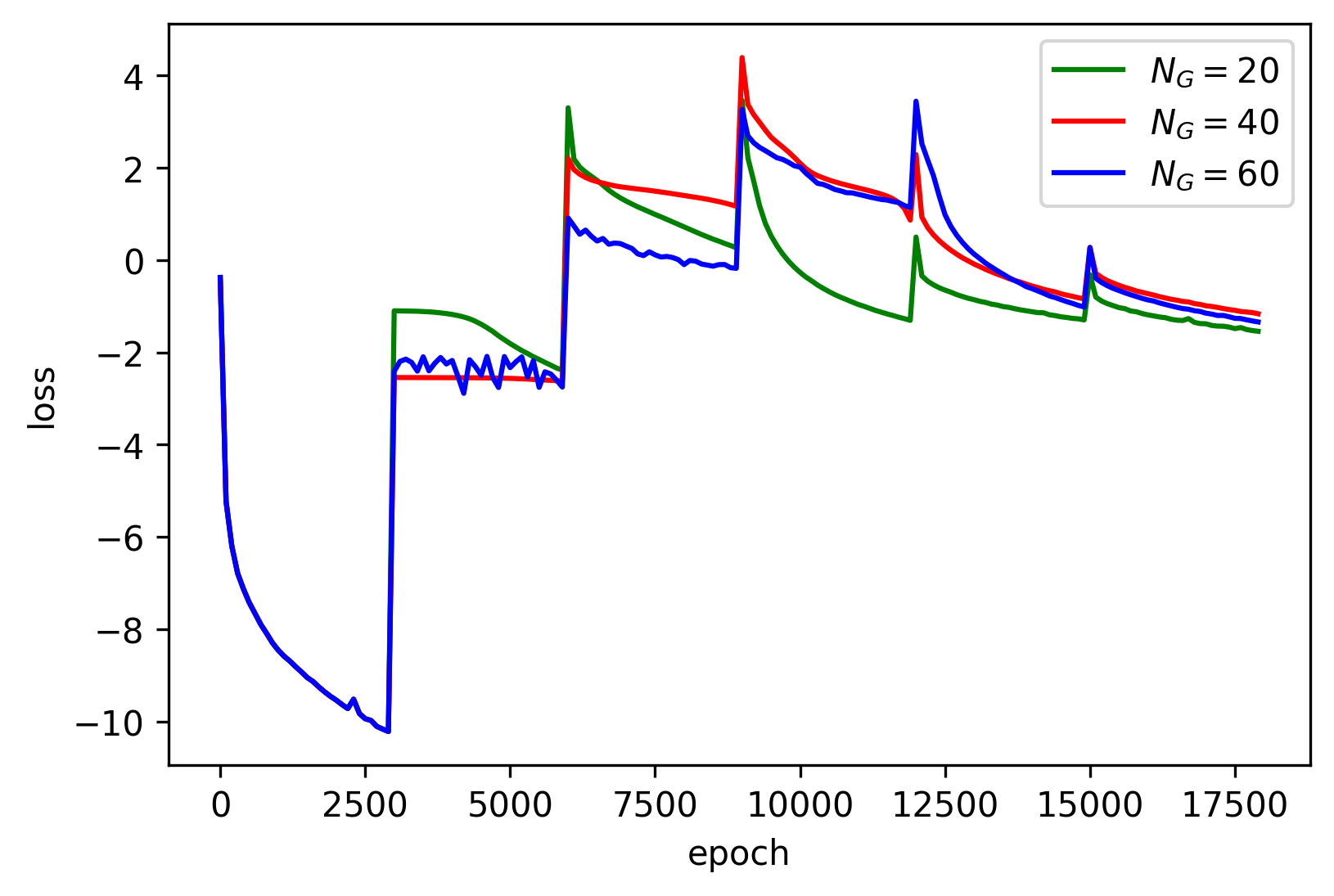}
}
\quad
\subfigure[]{
\includegraphics[scale=0.46]{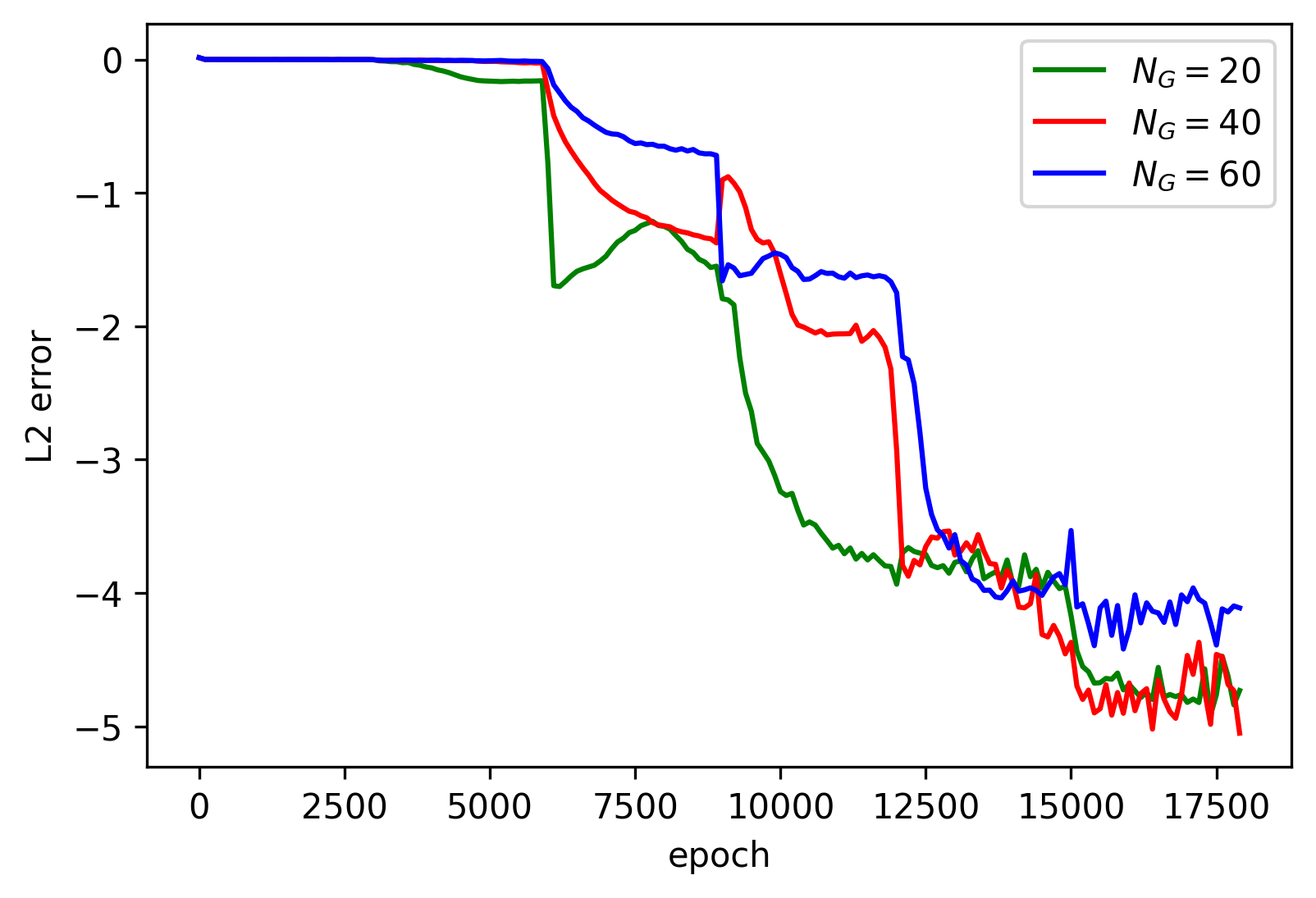}
}
\caption{The loss (a) and relative $L_2$ error (b) of GAS for different $N_G$ in ten dimension case.}
\label{fig:Compare_NG}     
\end{figure}

To visualize the adaptive training process in GAS, the loss and $L_2$ error with different $N_G$ are also depicted in Figure \ref{fig:Compare_NG}. As we can see, the $L_2$ error converges to $O(1\times 10^{-5})$ after 5 times of adaptive sampling. Before the first time GAS, the loss would converges to a rather small level while the $L_2$ error is almost unchanged, this reflects that the training of PINNs with a uniform distributed dataset would be misled to a trivial false solution, since statistically the points around singularity is rarely sampled. The following periodic jump in loss stands for the adjustment of risk introduced by the newly added points in GAS. Numerically, once the number of Gaussians in the mixture model reaches some value ($N_G = 20$ in this case), the increasing of $N_G$ would bring little gains in the $L_2$ error.

\section{Conclusion}
In this paper, we developed a Gaussian mixture distribution-based adaptive sampling (GAS) method and applied it into the training of physics-informed neural networks. The key idea of GAS is to construct a Gaussian mixture model to generate additional and important knowledge during the incremental learning. By using the coordinates and gradients of points with larger residual to define the means and covariances, our method may actively localize the singular region of the solution and speed up the convergence of PINNs method. Compared to existing adaptive sampling methods, GAS is easier to be implemented and can achieve lower error with even less training data and computational cost. Nevertheless, a systematic and rigorous analysis may help us better understand the efficiency and restriction of GAS, and extending GAS to more complex problems (e.g., using a deep generative model to replace the mixture model or using GAS to deal with problems with defects beyond point singularity) may further increase the power of PINNs in solving high dimensional singular problems. We will leave these topics for future study.

\appendix


\bibliographystyle{elsarticle-num-names}
\bibliography{ref}





\end{document}